\pdfoutput=1
\documentclass[11pt]{article}
\usepackage[preprint]{acl}
\usepackage{times}
\usepackage{latexsym}
\usepackage[T1]{fontenc}
\usepackage[utf8]{inputenc}
\usepackage{microtype}
\usepackage{graphicx}
\usepackage{amsmath,amssymb,amsthm}
\usepackage{booktabs}
\usepackage{multirow}
\usepackage{tikz}
\usetikzlibrary{arrows.meta,positioning,calc}

\newtheorem{theorem}{Theorem}
\newtheorem{proposition}[theorem]{Proposition}

\newtheorem{definition}[theorem]{Definition}
\theoremstyle{remark}
\newtheorem{remark}[theorem]{Remark}

\newcommand{\ASL}{\mathrm{ASL}}
\newcommand{\ASW}{\mathrm{ASW}}

\newcommand{\FRE}{\mathrm{FRE}}
\newcommand{\syl}{\mathrm{syl}}
\newcommand{\nat}{\,\mathrm{nat}}

\title{Interpreting Classical Readability Formulae as Surprisal}
\author{Yo Ehara \\
  Tokyo Gakugei University, Tokyo, Japan \\
  \texttt{ehara@u-gakugei.ac.jp}}

\begin{document}
\maketitle

\begin{abstract}
Surprisal correlates closely with reading time and is widely used in the cognitive science of reading as a measure of processing difficulty. Classical readability formulae such as the Flesch-Kincaid Grade Level, which scores a text by the US school grade expected to read it, are the most widely used practical measures of text difficulty, but their coefficients are unitless regression constants and have never been related to surprisal. We theoretically show that these coefficients can be interpreted as surprisal: every formula of the classical additive shape is a sum of per-word and per-sentence cross-entropies under a memoryless model, and once that model is tied to the language through Zipf's law and the formula is read as its linearisation at a reference text, each coefficient factorises into a scale in grades per nat, a Zipf slope and a reference sentence length. A grade thereby becomes a quantity in surprisal. Because the formula outputs grades and children's vocabulary has been measured by grade, the interpretation recovers the vocabulary growth the formula implicitly assumes and lets it be checked against development: holding sentence length fixed, the word coefficient of the Flesch-Kincaid formula converts to $0.18$ nat of unigram surprisal per word per grade ($0.15$ to $0.19$ across the regression-derived formulae), and normative root-word vocabulary growth (Biemiller and Slonim, 2001), converted through Zipf's law, gives $0.16$ nat per year at the frontier of the known vocabulary. The word coefficient is one year of vocabulary growth, and the grade is a vocabulary scale. The identification also yields a diagnostic (the sentence-length terms are not those of a memoryless surprisal) and extends to a Japanese formula.
\end{abstract}

\begin{figure}[t]
\centering
\begin{tikzpicture}[font=\small, >=Latex,
  box/.style={draw=black!60, rounded corners=2pt, inner sep=5pt, align=left, text width=\columnwidth-14pt},
  lab/.style={font=\footnotesize, align=left}]
\node[box, fill=blue!7] (form) {\textbf{Readability formulae} (education, plain language)\\[1pt]
$\mathrm{FKGL}=0.39\,\mathrm{ASL}+11.8\,\mathrm{ASW}-15.59$ (ASL: words per sentence; ASW: syllables per word)\\[1pt]
Output: a US school grade. Coefficients: unitless regression constants. ``$+1$ is one grade'' is interpretable, but a grade has no unit that any other measurement shares.};
\node[box, fill=black!4, below=5mm of form] (bridge) {\textbf{This paper}: Theorem 1 (every additive formula is a cross-entropy) $+$ Zipf identification ($-\log\pi(w)$ affine in syllables)\\[1pt]
$\Rightarrow$ coefficient $=$ (grades per nat) $\times$ (Zipf slope) $\times$ (sentence length)\\[1pt]
$\Rightarrow$ one grade $=0.15$ to $0.19$ nat of surprisal per word};
\node[box, fill=orange!9, below=5mm of bridge] (cog) {\textbf{Cognitive science of reading}\\[1pt]
Surprisal $-\log P(w)$ in nats or bits: reading time per bit in adults and children, eye movements by grade, vocabulary growth in root words per year (Zipf: $0.16$ nat per year).};
\draw[->, thick] (form) -- (bridge);
\draw[<->, thick] (bridge) -- (cog);
\end{tikzpicture}
\caption{Why read the coefficients as surprisal. The formulae and the cognitive science of reading measure text difficulty in incommensurable units; the surprisal reading converts the formulae' coefficients into nats, so that the Flesch-Kincaid grade step can be set against developmental measurements: one grade turns out to be the size of one year of vocabulary growth (Section~\ref{sec:dev}).}
\label{fig:bridge}
\end{figure}
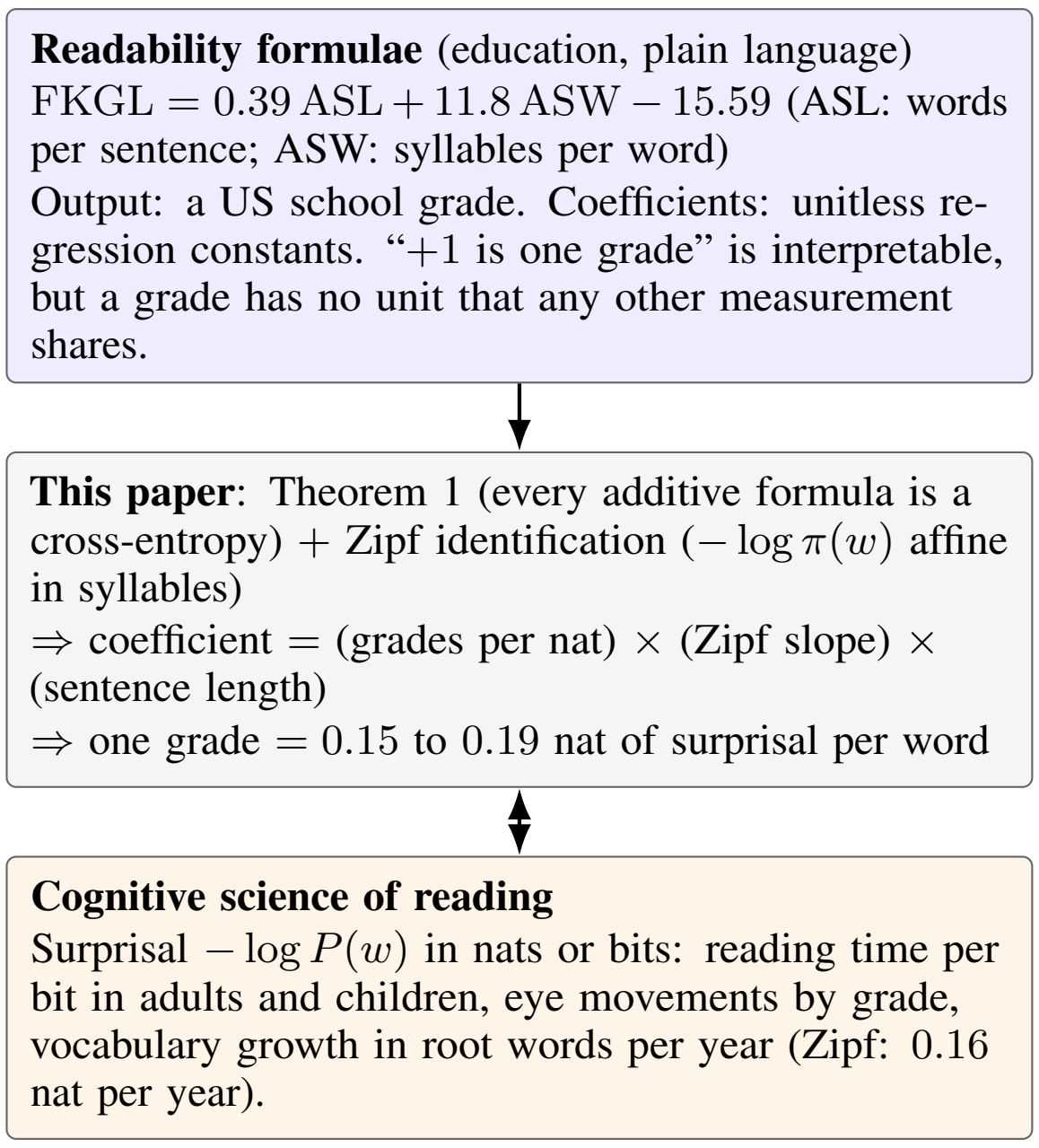

\section{Introduction}

Surprisal, the negative log probability of a word in context, is the standard currency of the cognitive science of reading: it predicts reading times with a linking function that has been confirmed across corpora and languages \citep{hale2001,levy2008,smith2013,shain2024,wilcox2023}, the frequency and context effects behind it have been traced through childhood \citep{joseph2013,tiffin2015,tiffin2020,reichle2013}, and children's vocabulary growth, reported in counts of words known per grade \citep{biemiller2001,anglin1993}, converts to the same log-frequency units. Mean surprisal has even been proposed as a readability measure in its own right \citep{gruteke2026}. Results in this literature are stated in bits or in nats and can be compared with one another. A nat is the unit of information when the logarithm is natural rather than binary, so $1$ nat $=1/\ln 2=1.443$ bits \citep{shannon1948,cover2006}; the choice is a constant factor, and both units are current in the field. Reading-time studies often report surprisal in bits \citep{smith2013,oh2023,gruteke2026}, whereas quantities that come from likelihoods, the log-probabilities of language models and the gains in regression log-likelihood by which predictors of reading time are compared \citep{meister2021}, are natural logarithms and so are in nats. Nats are also the native unit of the exponential-family and maximum-likelihood apparatus used here (a Gibbs distribution $e^{-\beta f}$, a regression of $-\log\pi$ on word length, Cram\'er's theorem), which would otherwise carry factors of $\ln 2$; we therefore derive in nats and give bits alongside the headline figures.

Readability formulae are the other tradition. The Flesch-Kincaid Grade Level \citep{kincaid1975}, the best known, scores a text as $0.39\times$ (words per sentence) $+\,11.8\times$ (syllables per word) $-\,15.59$, and the result is read as the US school grade expected to read the text: a score of 8 means eighth grade. It and its relatives (Flesch Reading Ease, the Automated Readability Index, Coleman-Liau, Gunning Fog, Dale-Chall, FORCAST) remain the most widely used practical measures of text difficulty, in classrooms, in plain-language law, and now as automatic evaluation signals for text simplification with language models. Almost all share one shape, a linear combination of average sentence length and the average of a per-word surface feature (syllables, letters, or membership in a list of familiar words), and their coefficients were obtained by regressing a comprehension criterion on those features \citep{flesch1948,kincaid1975,dale1948,coleman1975}. The coefficients are reported to two or three decimals and have been treated, ever since, as unexplained constants. A rise of one in the Flesch-Kincaid score is one grade, which is why the formula is so interpretable; but nobody can say what a grade is in the units in which reading is now studied, and the two traditions have hardly been put in contact. Language models have been used to \emph{replace} the formulae \citep{si2001,collins2004,martinc2021}, and the formulae have been criticised for having no theory \citep{howcroft2017,crossley2017}, but the question of what their coefficients would mean if the formulae were surprisal has not, to our knowledge, been asked.

This paper makes two claims, summarised in Figure~\ref{fig:bridge}.

\paragraph{The coefficients of the classical formulae can be interpreted as surprisal.} The first step is easy and, on its own, empty: any index that adds up a per-word feature is exactly a cross-entropy under some distribution, namely the Gibbs distribution that makes the feature a log-probability, and the sentence-length term is the surprisal of where sentences end under a geometric sentence-length process (Theorem~\ref{thm:rep}). Since this holds for every additive index whatsoever, it says nothing until the distribution is required to be one we can measure: the word-frequency distribution of the language. For syllable and letter counts that requirement is Zipf's law of abbreviation \citep{zipf1935,piantadosi2011}, in the form ``the surprisal of a word is affine in its length'', with a slope that frequency data determine. Under that identification, and reading the formula as the tangent plane (the linear approximation at a reference text) of a memoryless surprisal, each coefficient factorises into measurable quantities (Proposition~\ref{prop:coef}): the word coefficient is (grades per nat) $\times$ (nats per syllable, the Zipf slope) $\times$ (words per sentence), and the sentence-length coefficient is (grades per nat) $\times$ (nats per word, the unigram entropy of the language). Each factor is a fact about the language or about readers, so the identities are conditions that can be checked, and the checks are a diagnostic. Applied to English, the word terms of the four regression-derived grade formulae agree on about three nats of lexical surprisal per grade per sentence; their length terms are roughly four to eight times too light for any of them to be the tangent plane of a memoryless surprisal; and the sentence-length distribution the length term presupposes is not the one observed. Applied to Japanese, two of the four lexical coefficients of jReadability \citep{hasebe2015}, a regression-derived formula for learner texts, have the wrong sign to be the surprisal of Japanese.

\paragraph{The word coefficient is one year of vocabulary growth.} The Flesch-Kincaid formula outputs grades. Cognitive science has, independently, measured how children's vocabulary grows by grade and how the lexical variables behind surprisal govern their reading. Once a grade is a quantity in surprisal, the vocabulary growth the formula implicitly assumes can be recovered and set against the vocabulary growth that has been measured. Holding sentence length fixed, as a partial derivative does, the word coefficient of the Flesch-Kincaid formula converts to $0.18$ nat ($0.26$ bits) of unigram surprisal per word per grade ($0.15$ to $0.19$ nat across the four regression-derived grade formulae), a $2$ to $3\%$ increase on the per-word mean: the formula assumes that the words a child is expected to handle fall in frequency by about $15\%$ per grade. The normative root-word counts of \citet{biemiller2001}, converted through Zipf's rank-frequency law, give a rise of $0.16$ nat per year in the log-frequency frontier of the known vocabulary. The two numbers are the same. The word coefficient of the formula is therefore one year of vocabulary growth, written in syllables per word, and the formula's grade is a vocabulary scale: oral reading rate, by contrast, grows by about $10\%$ per grade \citep{hasbrouck2017}, several times faster on the scale on which surprisal predicts reading time, so the grade does not track fluency. The step is small because frequency is logarithmic in vocabulary size. This comparison could not be made while a grade was a number without units, and it is the reason to make the interpretation.

Beyond these two claims, the identification yields results that are useful on their own: which surface features are good linear predictors of surprisal (letters and the Dale-Chall list beat syllables), why the binary mono-/polysyllable split works, a Legendre-duality reading of why the formulae are linear and what the linearity costs, and two predictions for developmental surprisal studies that have not yet been run.

Section~\ref{sec:background} recalls the formulae and fixes units. Section~\ref{sec:rep} proves the representation theorem and says why it is empty. Section~\ref{sec:model} introduces the memoryless model and derives the coefficient identities. Section~\ref{sec:data} estimates the constants. Section~\ref{sec:read} reads the English formulae, Section~\ref{sec:dev} compares the grade step with developmental data, Section~\ref{sec:length} tests the length term, Section~\ref{sec:ja} treats jReadability, and Section~\ref{sec:disc} discusses what the interpretation licenses and what it does not.

\section{Background and units}
\label{sec:background}

\subsection{The formulae}
For a text $T$ with $N_s$ sentences, $N_w$ words and $N_y$ syllables, write $\ASL=N_w/N_s$ and $\ASW=N_y/N_w$. The English formulae considered here are listed in Table~\ref{tab:formulae}. All but two have the \emph{additive form}
\begin{equation}
F(T)=c_0+c_1\,\ASL+\frac{1}{N_w}\sum_{i=1}^{N_w} f(w_i),
\label{eq:additive}
\end{equation}
with a per-word feature $f$: syllable count ($f=b\,\syl$), letter count, an indicator of ``long'' (polysyllabic, or not on a familiar-word list) words, or a linear combination of several indicators. The two exceptions are SMOG, whose square root puts it outside the additive class, and Coleman-Liau, whose length term is $-29.6/\ASL$ rather than a multiple of $\ASL$; Coleman-Liau is handled below by local linearisation of its length term (Proposition~\ref{prop:coef}), not by Theorem~\ref{thm:rep}. For an ease score such as $\FRE$ we work with its negative so that $F$ increases with difficulty. The same form recurs outside English; Section~\ref{sec:ja} applies the analysis to jReadability, a Japanese formula whose $f$ is a linear combination of word-origin and part-of-speech indicators.

\begin{table*}[t]
\centering\footnotesize\setlength{\tabcolsep}{4pt}
\caption{Classical formulae. ``poly$_k$'' is the fraction of words with at least $k$ syllables (Gunning's ``complex words'' further exclude proper nouns, compounds and some suffixes, which we ignore); ``notDC'' the fraction of words not on the Dale-Chall familiar list; ``long$_6$'' the fraction of words longer than six letters; ``mono'' the number of one-syllable words in a 150-word sample. Coefficients are those of the original publications; the Navy recalculations are from \citet{kincaid1975}. All but SMOG and Coleman-Liau have the additive form \eqref{eq:additive}.}
\label{tab:formulae}
\begin{tabular}{llll}
\toprule
Formula & Scale & Word term & Length term \\
\midrule
Flesch Reading Ease \citep{flesch1948} & ease (RE) & $-84.6\,\ASW$ & $-1.015\,\ASL$ \\
Flesch-Kincaid Grade Level \citep{kincaid1975} & US grade & $11.8\,\ASW$ & $0.39\,\ASL$ \\
ARI \citep{smith1967} & US grade & $4.71\,\text{strokes/word}$ & $0.50\,\ASL$ \\
ARI, Navy recalculation \citep{kincaid1975} & US grade & $5.84\,\text{strokes/word}$ & $0.37\,\ASL$ \\
Coleman-Liau \citep{coleman1975} & US grade & $5.88\,\text{letters/word}$ & $-29.6/\ASL$ \\
Gunning Fog \citep{gunning1952} & US grade & $40\,\text{poly}_3$ & $0.4\,\ASL$ \\
Dale-Chall \citep{dale1948} & raw score & $15.79\,\text{notDC}$ & $0.0496\,\ASL$ \\
FORCAST \citep{caylor1973} & US grade & $-0.11\,\text{mono}$ & \\
FORCAST, Navy \citep{kincaid1975} & US grade & $-0.24\,\text{mono}$ & \\
Farr-Jenkins-Paterson, Navy \citep{kincaid1975} & US grade & $-0.307\,\%\text{mono}$ & $0.387\,\ASL$ \\
LIX \citep{bjornsson1968} & LIX & $100\,\text{long}_6$ & $1\,\ASL$ \\
SMOG \citep{mclaughlin1969} & US grade & \multicolumn{2}{l}{$1.043\sqrt{30\,N_{\text{poly}_3}/N_s}+3.13$ (not additive)} \\
\bottomrule
\end{tabular}
\end{table*}

\subsection{Surprisal and its units}
The surprisal of an event of probability $p$ is $I=-\log p$ \citep{shannon1948,cover2006}. With the natural logarithm the unit is the nat; with base 2 it is the bit; $1\nat=1/\ln 2=1.443$ bits. A bit is the information in one fair binary choice, so a word of surprisal $10.5$ bits is as unexpected as a particular outcome of $10.5$ coin flips, i.e.\ has probability about $1/1500$; the token-weighted mean surprisal of an English word under its unigram distribution is in fact $7.31\nat=10.54$ bits (Section~\ref{sec:data}), in line with Shannon's classical estimate of $11.8$ bits for a Zipfian vocabulary \citep{shannon1951}. Surprisal adds over independent events, so the surprisal of a sentence under a model in which words are independent is the sum of the words' surprisals; a statement such as ``one grade equals $3.2\nat$ per sentence'' means that the geometric-mean probability that the model assigns to a sentence falls by a factor $e^{3.2}\approx 24$ per grade, or that the reader resolves $4.6$ more binary alternatives per sentence. The mean surprisal of a sample under a model is the sample cross-entropy; its minimum over models is the entropy.

Surprisal theory links these quantities to behaviour: reading time increases linearly with the surprisal of a word in context \citep{smith2013,shain2024}, so a nat has an empirical price in milliseconds. We use that link only qualitatively; the identifications below are in nats, and the reader may convert.

\section{Every additive index is a cross-entropy}
\label{sec:rep}

Let $V$ be a finite vocabulary and $f:V\to\mathbb{R}$ a feature. For $\beta>0$ define the Gibbs (maximum-entropy) distribution with sufficient statistic $f$ \citep{jaynes1957},
\begin{equation}
\pi_{f,\beta}(w)=\frac{e^{-\beta f(w)}}{Z_\beta},\quad Z_\beta=\sum_{v\in V}e^{-\beta f(v)},
\label{eq:gibbs}
\end{equation}
so that $-\log\pi_{f,\beta}(w)=\beta f(w)+\log Z_\beta$ is affine in the feature with slope $\beta$. For a text $T=(w_1,\dots,w_{N_w})$ write the per-word cross-entropy $H_T(\pi)=N_w^{-1}\sum_i -\log\pi(w_i)$. For a boundary probability $q\in(0,1)$ let the \emph{geometric boundary process} decide, after each word and independently of the past, whether the sentence ends there, with probability $q$; the surprisal that a sentence of $L$ words ends exactly after its $L$th word is $(L-1)\,h_c+h_\#$ with $h_c=-\log(1-q)$ and $h_\#=-\log q$, and its average over the sentences of $T$ is
\begin{equation}
B_q(T)=h_c\,\ASL+(h_\#-h_c).
\label{eq:boundary}
\end{equation}

\begin{theorem}[Representation]
\label{thm:rep}
Let $F$ have the additive form \eqref{eq:additive}. For every $\beta>0$ and $q\in(0,1)$,
\begin{align*}
F(T)={}&\frac{1}{\beta}\,H_T(\pi_{f,\beta})+\frac{c_1}{h_c}\,B_q(T)\\
&+\Bigl(c_0-\frac{\log Z_\beta}{\beta}-\frac{c_1 (h_\#-h_c)}{h_c}\Bigr).
\end{align*}
If $f=\boldsymbol\beta^{\top}\boldsymbol\phi$ is a linear combination of several features, the same holds with $\pi_{f,\beta}$ the log-linear model $\pi(w)\propto\exp(-\beta\,\boldsymbol\beta^{\top}\boldsymbol\phi(w))$, whose regression coefficients are the model's weights.
\end{theorem}
\begin{proof}
By \eqref{eq:gibbs}, $\beta^{-1}H_T(\pi_{f,\beta})=N_w^{-1}\sum_i f(w_i)+\beta^{-1}\log Z_\beta$. By \eqref{eq:boundary}, $c_1\ASL=(c_1/h_c)\bigl(B_q(T)-(h_\#-h_c)\bigr)$. Add.
\end{proof}

Theorem~\ref{thm:rep} is exact and holds for every additive index, which is precisely why it is empty: the distribution $\pi_{f,\beta}$ is manufactured from the formula. Three further facts are needed before the representation says anything about reading.

\begin{enumerate}
\item \emph{Identification of the slope.} The induced $\pi_{f,\beta}$ must be related to a distribution with independent meaning: the unigram distribution $\pi$ of the language, or the predictive distribution $\pi_r$ of a reader. What the coefficient identities of Section~\ref{sec:model} need is not that $\pi_{f,\beta}$ be close to $\pi$ in Kullback-Leibler divergence (for a bounded feature such as the syllable count the Gibbs model spreads its mass uniformly within each feature class and is far from any natural language in that sense) but that the feature predict $-\log\pi(w)$ well as an affine function over tokens, with a slope that the data determine. We therefore measure the identification by the token-weighted least-squares fit of $-\log\pi(w)$ on $f(w)$, report its slope and its $R^2$, and use the slope in the identities; the fitted intercept is free, unlike the intercept $\log Z_\beta$ of the Gibbs model, so this is an identification of the slope, not of the distribution. For syllable and letter counts this identification \emph{is} Zipf's law of abbreviation in its logarithmic form \citep{zipf1935,sigurd2004,piantadosi2011,levshina2022}.
\item \emph{Consistency across features.} If several features are all to be surprisal under the same $\pi$, they must be affinely related to one another on the token distribution; their residuals from one another measure what each captures beyond surprisal.
\item \emph{Accounting unit.} Theorem~\ref{thm:rep} mixes a per-word unit (the cross-entropy) with a per-sentence unit (the boundary term). A single generative model yields one unit, and the next section shows which unit the formulae' signs require within such a model; the classical formulae are then compared with the model's tangent plane.
\end{enumerate}

\begin{remark}[Legendre duality]
\label{rem:legendre}
Theorem~\ref{thm:rep} has a geometric reading that explains why classical formulae are linear and what the linearity costs. The family \eqref{eq:gibbs} is an exponential family with natural parameter $\beta$ and log-partition function $A(\beta)=\log Z_\beta$; the feature mean $m_T=N_w^{-1}\sum_i f(w_i)$ is its mean parameter, and the two are Legendre dual \citep{wainwright2008}. The cross-entropy of a text under a member with \emph{fixed} $\beta$ is linear in the mean parameter, $H_T(\pi_{f,\beta})=\beta m_T+A(\beta)$, and this linear function is a supporting hyperplane of the concave entropy function of the family in mean coordinates,
\begin{gather*}
S(m)=\min_{\beta}\bigl[\beta m+A(\beta)\bigr],\\ S'(m)=\beta(m),\qquad S''(m)=-\frac{1}{\mathrm{Var}_{\beta(m)}(f)},
\end{gather*}
which is the entropy of the member whose feature mean is $m$, i.e.\ of the model re-fitted to the text. The gap between the two is a Bregman divergence,
\begin{align*}
\beta_0 m+A(\beta_0)-S(m)&=D_{\mathrm{KL}}\bigl(\pi_{\beta(m)}\,\|\,\pi_{\beta_0}\bigr)\\&\approx\frac{(m-m_0)^2}{2\,\mathrm{Var}_{\beta_0}(f)},
\end{align*}
where $m_0$ is the feature mean at which $\beta(m_0)=\beta_0$. A classical formula therefore fixes the slope $\beta$ (the reader's inverse temperature) and does not re-estimate it text by text: it is the tangent of $S$ at the calibration point, exact there and in error by a term that is quadratic in the distance from the calibration point and inversely proportional to the variance of the feature. The same holds for the length term, because the geometric distribution is the exponential family on $\{1,2,\dots\}$ with sufficient statistic $L$, natural parameter $-h_c$ and mean parameter $L_0=1/q$; so both terms of a formula are cross-entropies at fixed natural parameters, written in the mean coordinates $(\ASL,\ASW)$, and the Legendre transform of the family's log-partition function (not of a single formula, whose convex conjugate is degenerate) gives the nonlinear, re-fitted counterpart $S(\ASL,\ASW)$ of which each formula is one supporting plane, for feature means inside the range the family can realise. In thermodynamic terms, $m$ is an energy, $1/\beta$ a temperature and $-\beta^{-1}\log Z_\beta$ a free energy, and Theorem~\ref{thm:rep} is the identity $U=F+TS$ with the cross-entropy in the place of the entropy. Two cautions. First, the gap is measured against the re-fitted entropy $S$, not against the fixed-$\pi$ model $\bar I_0$ of Section~\ref{sec:model}, which is exactly linear in $\ASW$ and whose own linearisation error has the two components computed in Appendix~\ref{app:deriv}; the two gaps answer different questions (does the reader re-adapt to each text, or not) and are not to be added. Second, the variance in the denominator is the variance of $f$ under the family member at the calibration point, which for the maximum-entropy family \eqref{eq:gibbs} is a function of $\beta_0$ and the type counts, not of the language. It equals the language variance only in the language-anchored family of Section~\ref{sec:disc}, with the calibration point at the language mean, where the gap is the Cram\'er rate function. With that family and $f=\syl$, $\mathrm{Var}_\pi(\syl)=0.66$ (Section~\ref{sec:data}) gives, for a text whose $\ASW$ differs from the language mean by $0.2$, a gap of about $0.03\nat$ per word, or $0.5\nat$ per sentence at $L_0=17.6$ words, i.e.\ $0.17$ grades at $\kappa=0.315$ grades per nat (Section~\ref{sec:read}); a difference of $0.4$ gives $0.67$ grades. These numbers are the model-free surprisal of a text's syllable profile relative to the language, not the cost of linearity under the Zipf reading of Sections~\ref{sec:model} to \ref{sec:read}: under the maximum-entropy family at $\beta_0=\lambda=2.13$, the variance of the syllable count over our $97{,}724$ types is $0.17$, so the corresponding gap is about four times larger ($0.65$ grades at a distance of $0.2$), and that member has a mean syllable count of $1.18$ rather than the language's $1.47$; real texts, with $\ASW$ between $1.4$ and $1.6$, sit $0.2$ to $0.4$ from that member's mean, where the gap is $0.8$ to $2.9$ grades, a reminder that the Zipf identification fixes the slope of the surprisal and not the distribution.
\end{remark}

The theorem also fixes what is \emph{not} a surprisal. Thresholded features (an indicator that a word has three or more syllables, or is absent from a list) induce two-point surprisal values and can be surprisal under $\pi$ only to the extent that $-\log\pi$ is close to a step function of the feature; the square root in SMOG puts it outside the additive class altogether.

\section{The memoryless model and the coefficient identities}
\label{sec:model}

\begin{definition}[Memoryless text model]
A memoryless model $\mathcal M_0=(\pi,q)$ generates a sentence by emitting a word $w\sim\pi$ and then, independently of the past, ending the sentence with probability $q$ or emitting another word otherwise. Sentence lengths are therefore geometric, $P(L)=q(1-q)^{L-1}$ on $L\ge1$, the unique memoryless distribution on the positive integers; words are independent of one another and of their position.
\end{definition}

The surprisal of a sentence $s=(w_1,\dots,w_L)$ under $\mathcal M_0$ is $I_0(s)=\sum_{i\le L}-\log\pi(w_i)+(L-1)\,h_c+h_\#$, and its average over the sentences of $T$ is
\begin{equation}
\begin{aligned}
\bar I_0(T)&=\frac{1}{N_s}\sum_{s\in T}I_0(s)\\&=\ASL\cdot\bar\imath_T+h_c\,\ASL+(h_\#-h_c),\\
\bar\imath_T&\equiv N_w^{-1}\textstyle\sum_i-\log\pi(w_i).
\end{aligned}
\label{eq:Ibar}
\end{equation}
Within this model the sign of the length term fixes the accounting unit: per word, the boundary surprisal is $h_c+(h_\#-h_c)/L$, which decreases with sentence length whenever $q<1/2$, whereas every formula in Table~\ref{tab:formulae} that has a length term charges more for longer sentences. We therefore account per sentence. The unit is a convention of the model, and where it matters below we report per-word values as well.

Now identify $\pi$ with the language and invoke the logarithmic Zipf law, $-\log\pi(w)\approx\lambda\,\syl(w)+\mu$ in token-weighted least squares (the weighting matters: $\ASW$ is a token mean, and we show in Section~\ref{sec:data} that the type-weighted slope on our list is essentially zero). Then $\bar\imath_T\approx\lambda\,\ASW+\mu$ and
\begin{equation}
\bar I_0(T)\approx\lambda\,\ASL\cdot\ASW+(\mu+h_c)\,\ASL+(h_\#-h_c).
\label{eq:bilinear}
\end{equation}
The leading term is $\lambda$ times the number of syllables per sentence, $N_y/N_s=\ASL\cdot\ASW$: under a memoryless Zipfian reader, the information in a sentence is its syllable count at $\lambda$ nats per syllable. The number of syllables per sentence has been identified, by algebraic rearrangement, as the quantity the Flesch-Kincaid formulae effectively measure \citep{ehara2024paclic}; \eqref{eq:bilinear} supplies a probabilistic reason.

The classical formulae are linear in $(\ASL,\ASW)$. Linearising \eqref{eq:bilinear} at a reference point $(L_0,A_0)$ and allowing an affine calibration $\mathrm{GL}=\kappa\bar I_0+\gamma$ to the grade scale gives:

\begin{proposition}[Coefficient identities]
\label{prop:coef}
Under the memoryless model with the logarithmic Zipf law, the coefficients of a formula $\mathrm{GL}=\beta_{\ASL}\ASL+\beta_{\ASW}\ASW+c$ that is the tangent plane of $\kappa\bar I_0+\gamma$ at $(L_0,A_0)$ satisfy
\begin{equation}
\begin{aligned}
\beta_{\ASW}&=\kappa\lambda L_0,\\ \beta_{\ASL}&=\kappa(\lambda A_0+\mu+h_c)=\kappa(H+h_c),
\end{aligned}
\label{eq:ident}
\end{equation}
where $H=\lambda A_0+\mu$ is the token-mean unigram surprisal (the cross-entropy of the language under $\pi$) because weighted least squares passes through the weighted means. Consequently the ratio of the two coefficients is predicted without $\kappa$,
\begin{equation}
\frac{\beta_{\ASW}}{\beta_{\ASL}}=\frac{\lambda L_0}{H+h_c},
\label{eq:ratio}
\end{equation}
and the scale $\kappa$ can be estimated twice, from either coefficient:
\begin{equation}
\kappa_{\mathrm{lex}}=\frac{\beta_{\ASW}}{\lambda L_0},\qquad \kappa_{\mathrm{len}}=\frac{\beta_{\ASL}}{H+h_c}.
\label{eq:twokappa}
\end{equation}
\end{proposition}

For a letter-count formula, replace $\lambda$ by the letter slope (we use it also for the typewriter-stroke count of the two ARIs, which adds digits and punctuation to the letters); for an indicator feature $\mathbf 1_C$ of a word class $C$, the Zipf law is replaced by the two-class approximation $-\log\pi(w)\approx\delta_C\mathbf 1_C(w)+\text{const}$, where $\delta_C$ is the token-weighted difference in mean surprisal between $C$ and its complement, and $\beta_C=\kappa\,\delta_C L_0$ for a coefficient on the \emph{fraction} of class-$C$ words; for several indicators used together, $\delta$ is the vector of partial coefficients of the token-weighted regression of $-\log\pi$ on the indicators (Section~\ref{sec:ja}). Equation \eqref{eq:twokappa} is the crux: the model has one free scale, and a formula whose two coefficients return different scales is not the tangent plane of a memoryless surprisal.

\paragraph{What the identities assume.} Proposition~\ref{prop:coef} equates a regression coefficient with a partial derivative. That is a structural reading, and three of its assumptions deserve to be named, because a failure of \eqref{eq:twokappa} can be a failure of any of them. (i) \emph{Tangent plane versus global fit.} The coefficients of a formula fitted by least squares on a finite set of passages are not the derivatives of any function at a point; even if the criterion were exactly $\kappa\bar I_0+\gamma$ with $\bar I_0$ bilinear as in \eqref{eq:bilinear}, the fitted coefficients would differ from \eqref{eq:ident} by terms that depend on the joint distribution of $(\ASL,\ASW)$ in the calibration sample. (ii) \emph{Reference point.} $\kappa_{\mathrm{lex}}$ scales as $1/L_0$, and the typical sentence length of the calibration passages (school readers for ARI, Navy manuals for FKGL) is not known to equal the reference we use. (iii) \emph{Proxy and aggregation.} The Zipf fit is a token-level regression; its slope need not equal the slope of mean surprisal on $\ASW$ across texts, which is what a criterion regression across passages would see. Section~\ref{sec:data} measures the third assumption directly; the first two are discussed with the results.

\section{Estimating the constants}
\label{sec:data}

\paragraph{Word distribution.} We take $\pi$ from version 3.1.1 of the \texttt{wordfreq} frequency lists \citep[list \texttt{best}]{speer2022}, keeping the $97{,}724$ alphabetic types among the top $100{,}000$ and renormalising (they carry $95.7\%$ of the list's mass). Syllables are counted from CMUdict \citep{cmudict} with \texttt{textstat}'s heuristic as fallback \citep{textstat}; letters are characters. The Dale-Chall familiar-word list is the one shipped with \texttt{textstat} ($2{,}941$ entries). All means and regressions are token-weighted by $\pi$ unless stated. When texts are scored (below), words outside the list are given the surprisal of a frequency of $10^{-9}$.

\paragraph{Sentence lengths and a second frequency source.} The Brown corpus \citep{francis1979} supplies $56{,}829$ sentences for the length distribution and a second unigram distribution for a robustness check of the Zipf slope (types with at least two tokens, $98.3\%$ of tokens, renormalised). Its $500$ documents also serve to measure the text-level slope across genres.

\paragraph{A levelled corpus.} OneStopEnglish \citep{vajjala2018} provides 189 news articles each rewritten at three levels (elementary, intermediate, advanced; $567$ texts). We compute $\ASL$, $\ASW$, the Flesch-Kincaid scores and the per-word and per-sentence unigram surprisal of each text, with sentences split by the Punkt tokenizer; the three versions of an article are not independent, and we use them only descriptively.

\paragraph{Japanese.} For jReadability we take the \texttt{wordfreq} Japanese list (top $150{,}000$ entries), tag each entry with UniDic \citep{den2007} through \texttt{fugashi} \citep{mccann2020}, keep the $110{,}049$ entries that are a single UniDic short unit ($96.6\%$ of the list's mass) and record their word origin (\emph{goshu}) and part of speech.

\begin{table*}[t]
\centering\footnotesize
\caption{English constants (token-weighted unless marked). $H$: unigram cross-entropy per word; $\lambda$, $\mu$, $R^2$: token-weighted least squares of $-\log\pi(w)$ on the feature; $\delta_C$: surprisal difference between class $C$ and its complement; $\tau^\ast$: the surprisal threshold that best reproduces class membership, with its token-weighted balanced accuracy. The Brown regression uses types with at least two tokens, for which the mean syllable count is $1.520$.}
\label{tab:constants}
\begin{tabular}{lll}
\toprule
 & wordfreq & Brown \\
\midrule
$H$ (nat / bit per word) & $7.31$ / $10.54$ & $7.10$ / $10.24$ \\
mean syllables per word $A_0$ (all tokens) & $1.474$ & $1.537$ \\
variance of the syllable count per word & $0.66$ & \\
mean letters per word & $4.46$ & \\
$\lambda_{\syl}$ (nat per syllable), $\mu$, $R^2$ & $2.13$, $4.17$, $0.34$ & $2.00$, $4.06$, $0.36$ \\
$\lambda_{\syl}$ type-weighted, $R^2$ & $-0.01$, $0.00$ & \\
$\lambda_{\mathrm{letter}}$ (nat per letter), $\mu$, $R^2$ & $0.90$, $3.31$, $0.52$ & \\
\midrule
\multicolumn{3}{l}{\emph{Mean surprisal by syllable count} (nat; token mass in parentheses), wordfreq} \\
1 / 2 / 3 syllables & \multicolumn{2}{l}{$6.04\,(.68)$ / $9.67\,(.20)$ / $10.54\,(.08)$} \\
4 / 5 / 6 syllables & \multicolumn{2}{l}{$10.91\,(.03)$ / $11.32\,(.007)$ / $12.61\,(.0007)$} \\
\midrule
\multicolumn{3}{l}{\emph{Indicator features}: mass, $\delta_C$ (nat), $R^2$, $\tau^\ast$ (balanced accuracy), wordfreq} \\
poly$_2$ ($\ge2$ syllables) & \multicolumn{2}{l}{$0.318$, $4.00$, $0.39$, $7.55$ ($0.83$)} \\
poly$_3$ ($\ge3$ syllables) & \multicolumn{2}{l}{$0.114$, $3.81$, $0.17$, $7.80$ ($0.81$)} \\
long$_6$ ($>6$ letters) & \multicolumn{2}{l}{$0.194$, $4.02$, $0.29$, $7.80$ ($0.83$)} \\
notDC (not on Dale-Chall list) & \multicolumn{2}{l}{$0.285$, $4.82$, $0.53$, $8.30$ ($0.90$)} \\
\midrule
\multicolumn{3}{l}{\emph{Text-level slope of mean unigram surprisal on $\ASW$} (nat per syllable)} \\
OneStopEnglish, across its 567 texts & \multicolumn{2}{l}{$1.47$} \\
OneStopEnglish, within an article (advanced minus elementary) & \multicolumn{2}{l}{$1.97$ (least squares), $2.63$ (origin)} \\
Brown, across its 500 documents (15 genres) & \multicolumn{2}{l}{$0.20$ ($r=0.10$)} \\
sd of the text residual $\bar r_T$, OneStopEnglish (nat/word) & \multicolumn{2}{l}{$0.23$} \\
\midrule
\multicolumn{3}{l}{\emph{Sentence length}, Brown ($56{,}829$ sentences)} \\
mean / median / mode $L$ & \multicolumn{2}{l}{$17.62$ / $15$ / $11$} \\
variance of $L$: observed / geometric with the same mean & \multicolumn{2}{l}{$138.7$ / $292.8$} \\
$h_c$, $h_\#$ at $q=1/17.62$ (nat) & \multicolumn{2}{l}{$0.058$, $2.87$} \\
\bottomrule
\end{tabular}
\end{table*}

Table~\ref{tab:constants} collects the English constants. Five points matter for what follows.

\emph{The Zipf slope is a token-weighted quantity.} Over tokens, surprisal rises by $2.13\nat$ ($3.07$ bits) per syllable and $0.90\nat$ per letter; over the types of our list the slope vanishes, because the vast majority of types are rare and, within the top $100{,}000$, their rarity does not depend on their length (the classical, type-based statements of the law of abbreviation concern the full vocabulary and are not contradicted). The formulae average over tokens, so the token slope is the relevant one. The two frequency sources agree ($2.13$ vs.\ $2.00$ nat per syllable).

\emph{The token slope does not carry to texts in general.} Equation \eqref{eq:bilinear} substitutes a token-level regression into a text-level mean, which is legitimate only if the texts being compared differ in $\ASW$ for the reason the token fit captures. They often do not. Across the $567$ OneStopEnglish texts, all newspaper prose, the slope of mean unigram surprisal on $\ASW$ is $1.47\nat$ per syllable; within an article, comparing the advanced with the elementary version, it is $1.97$ (least squares), $2.63$ (a line through the origin) or $2.96$ (the ratio of the level-mean differences, Table~\ref{tab:ose}); the text-level slope depends on the estimator as well as on the genre. Across the $500$ documents of Brown, which span fifteen genres, it is $0.20$ with a correlation of $0.10$: genre moves mean surprisal independently of word length (technical prose uses long but, within its domain, frequent words). The text-specific residual of the token fit, $\bar r_T=\bar\imath_T-\lambda\,\ASW_T-\mu$, has a standard deviation of $0.23\nat$ per word across OneStopEnglish texts, which is $4\nat$ per sentence at $L_0$, or $1.3$ grades at the scale of Section~\ref{sec:read}. The identification of a word coefficient with $\kappa\lambda L_0$ is therefore meaningful within a genre, where the text-level slope brackets the token slope, and carries a systematic uncertainty of roughly a factor of $1.5$ even there.

\emph{The relation is concave.} Surprisal jumps by $3.6\nat$ from one to two syllables and then by only $0.9$, $0.4$ and $0.4$ (the token-weighted variance of the syllable count, $0.66$, enters the curvature term of Remark~\ref{rem:legendre}). A linear fit in the syllable count is therefore crude ($R^2=0.34$), and a single mono-/polysyllable indicator, used linearly, predicts surprisal better ($R^2=0.39$); the count carries at least as much information as the indicator, but the formulae use it linearly. Letters do better than syllables ($0.52$), and the Dale-Chall list, which was built from children's familiarity judgments rather than from length, is the best linear predictor of the four ($0.53$): non-membership in the list is reproduced with a token-weighted balanced accuracy of $0.90$ by the single rule $-\log\pi(w)>8.3\nat$, i.e.\ frequency below about $2.5$ per $10{,}000$ words. The familiar-word list is, to a good approximation, a frequency threshold.

\emph{Boundary surprisal is negligible per word.} With $q=1/17.62$, $h_c=0.058\nat$; the memoryless model attributes essentially all of the cost of a longer sentence to the words it contains ($H+h_c=7.37\nat$ per word), not to the boundary.

\emph{Sentence length is under-dispersed.} The geometric distribution with the Brown mean would have variance $292.8$; the observed variance is $138.7$. We return to this in Section~\ref{sec:length}.

\section{Reading the English formulae}
\label{sec:read}

\begin{table*}[t]
\centering\footnotesize\setlength{\tabcolsep}{4pt}
\caption{The coefficients of Table~\ref{tab:formulae} read through Proposition~\ref{prop:coef} with the constants of Table~\ref{tab:constants} and $L_0=17.62$ (Brown); all entries are computed from unrounded constants. $\beta_{\mathrm{lex}}$ is the word-term coefficient per unit of the feature; ``slope'' is $\lambda$ for counts and $\delta_C$ for indicators (nat per unit); $\kappa_{\mathrm{lex}}=\beta_{\mathrm{lex}}/(\text{slope}\cdot L_0)$ and $\kappa_{\mathrm{len}}=\beta_{\ASL}/(H+h_c)$ are in scale units per nat of per-sentence surprisal; ``1 unit'' is $1/\kappa_{\mathrm{lex}}$. For Coleman-Liau the length coefficient is the derivative of $-29.6/\ASL$ at $L_0$. For FORCAST, which has no length term and was designed for non-continuous military prose, the use of $L_0$ is a convention. Dale-Chall is on its raw-score scale, which its conversion table maps to roughly one to two grades per unit.}
\label{tab:kappa}
\begin{tabular}{lllrrrrrr}
\toprule
Formula & scale & feature & $\beta_{\mathrm{lex}}$ & $\kappa_{\mathrm{lex}}$ & 1 unit (nat / bit) & $\beta_{\ASL}$ & $\kappa_{\mathrm{len}}$ & ratio \\
\midrule
FRE & RE & syl/word & $84.6$ & $2.26$ & $0.44$ / $0.64$ & $1.015$ & $0.138$ & $16.4$ \\
FKGL & grade & syl/word & $11.8$ & $0.315$ & $3.17$ / $4.58$ & $0.39$ & $0.053$ & $6.0$ \\
ARI & grade & strokes/word & $4.71$ & $0.298$ & $3.35$ / $4.84$ & $0.50$ & $0.068$ & $4.4$ \\
ARI (Navy) & grade & strokes/word & $5.84$ & $0.370$ & $2.71$ / $3.90$ & $0.37$ & $0.050$ & $7.4$ \\
Coleman-Liau & grade & letters/word & $5.88$ & $0.372$ & $2.69$ / $3.88$ & $0.095$ & $0.013$ & $28.7$ \\
Gunning Fog & grade & poly$_3$ & $40$ & $0.595$ & $1.68$ / $2.42$ & $0.40$ & $0.054$ & $11.0$ \\
Dale-Chall & raw & notDC & $15.79$ & $0.186$ & $5.37$ / $7.75$ & $0.0496$ & $0.007$ & $27.6$ \\
FORCAST & grade & poly$_2$ & $16.5$ & $0.234$ & $4.27$ / $6.16$ & & & \\
FORCAST (Navy) & grade & poly$_2$ & $36.0$ & $0.511$ & $1.96$ / $2.82$ & & & \\
FJP (Navy) & grade & poly$_2$ & $30.7$ & $0.436$ & $2.30$ / $3.31$ & $0.387$ & $0.053$ & $8.3$ \\
LIX & LIX & long$_6$ & $100$ & $1.41$ & $0.71$ / $1.02$ & $1$ & $0.136$ & $10.4$ \\
\bottomrule
\end{tabular}
\end{table*}

Table~\ref{tab:kappa} applies the identities. Three regularities stand out.

\subsection{A conditional conversion: about three nats per grade per sentence}
The four grade-scale formulae that were fitted by regression on a continuous length proxy, namely FKGL (syllables, cloze-derived grade criterion on Navy manuals), ARI (typewriter strokes, graded school materials), the Navy ARI (strokes, the same Navy criterion) and Coleman-Liau (letters, cloze criterion), return $\kappa_{\mathrm{lex}}=0.32$, $0.30$, $0.37$ and $0.37$ grades per nat. Converted, one US school grade corresponds to $2.7$ to $3.4\nat$, or $3.9$ to $4.8$ bits, of lexical surprisal per sentence at a sentence length of $17.6$ words; per word this is $0.15$ to $0.19\nat$ ($0.22$ to $0.27$ bits) per grade. The unit-free content of the agreement is the ratio of the word coefficient to the slope of its feature: $11.8/2.13=5.5$, $4.71/0.90=5.2$, $5.84/0.90=6.5$ and $5.88/0.90=6.5$ grades per nat of mean per-word surprisal (the reciprocal is the step in nats per word per grade), within $\pm12\%$ of their mean, although the syllable and letter slopes differ by a factor of $2.4$. Three qualifications limit what this shows. The four calibrations are not four independent replications: FKGL and the Navy ARI were refitted in one study on the same eighteen passages \citep{kincaid1975}, and the original and Navy ARIs differ by $24\%$ in the same ratio although they use the same feature. The conversion is conditional on $L_0$ and on the token-level slope; with the text-level slopes of Section~\ref{sec:data} ($1.5$ to $3.0$ within a genre, depending on the estimator) the FKGL scale ranges from $0.22$ to $0.45$ grades per nat. And the formulae were historically adjusted to agree with one another on grade predictions, so agreement in $\kappa$ partly reflects agreement in the ratios of the features' slopes across texts. We therefore read the number as the conversion between grades and lexical nats that the formulae imply within the genre they were calibrated on, with an uncertainty of about a factor of $1.5$, and not as a measurement of what a year of schooling buys.

The indicator-based formulae spread more widely ($0.19$ to $0.60$), as expected from a two-point proxy for a graded quantity; Gunning Fog, which was never fitted by regression, is the outlier at the top, and the original FORCAST at the bottom. The Flesch Reading Ease scale comes out at $2.26$ RE points per nat, so $0.44\nat$ per point.

\subsection{The length term is four to eight times too light}
If the memoryless model were right and the formula were its tangent plane, the scale read off the length coefficient would equal the scale read off the word coefficient. It does not: $\kappa_{\mathrm{len}}$ is $0.05$ to $0.07$ grades per nat for every regression-derived grade formula with a linear length term (FKGL, both ARIs and the Navy FJP), against $\kappa_{\mathrm{lex}}=0.30$ to $0.44$ for those four (the Navy FJP, not Coleman-Liau, is the fourth here); Coleman-Liau, whose length term is $-29.6/\ASL$ and therefore has a slope that depends on $L_0$, is lighter still ($0.013$). The ratio in the last column of Table~\ref{tab:kappa} is $4$ to $8$ for the four, $16$ for Flesch Reading Ease and $29$ for Coleman-Liau, and the prediction \eqref{eq:ratio}, $\beta_{\ASW}/\beta_{\ASL}=\lambda L_0/(H+h_c)=5.1$, is contradicted by the observed $30.3$ (FKGL) and $83.4$ (FRE). The direction is the same for every formula, and the size is far outside the factor-of-$1.5$ allowance for the slope. Nor can the reference length rescue it: since $\kappa_{\mathrm{len}}$ does not depend on $L_0$ (up to the tiny $h_c$) while $\kappa_{\mathrm{lex}}$ scales as $1/L_0$, the two FKGL scales would coincide only at $L_0=11.8/(2.13\times0.053)\approx105$ words per sentence, and for any reference length up to $40$ words the mismatch remains a factor of about two or more. This is the robust result of the diagnostic: at any realistic reference text, no classical formula is the tangent plane of a memoryless surprisal of the language. The ratio itself is not a stable constant across formulae, which is itself informative: whatever makes the length term light is not a universal number.

Read the other way round, the formulae tell us what an additional word costs on their own scale: solving $\beta_{\ASL}=\kappa_{\mathrm{lex}}(\bar\imath_{\mathrm{eff}}+h_c)$ gives $\bar\imath_{\mathrm{eff}}=1.18\nat$ ($1.7$ bits) per word for FKGL and $0.39\nat$ ($0.56$ bits) for FRE, against $7.31\nat$ for a word drawn from the unigram distribution. The sentence-length coefficient charges an extra word at one-sixth of its unigram information. Several explanations are available, and the data of this paper do not separate them. One is context: a reader with memory pays $-\log P(w_i\mid w_{<i})$ rather than $-\log\pi(w_i)$, and the expected difference, the mutual information between a word and its context, accumulates over the sentence. Contextual language-model surprisal of English news averages about $5$ bits per word under a small model (Pythia-70M; \citealp{gruteke2026}), i.e.\ $3.5\nat$, and less under larger models; even so, the formulae' implied cost is lower by a further factor of three, so context alone does not close the gap. A second is that part of the length term is not lexical at all (an integration cost of the kind proposed by \citealp{gibson1998}, roughly constant per word). A third is the set of assumptions named after Proposition~\ref{prop:coef}: the coefficients of a global least-squares fit are not tangent-plane derivatives, the reference length of the calibration passages is unknown, and the criteria (cloze-derived grade levels for the Navy formulae and Coleman-Liau, comprehension tests or graded school materials for the older ones) may simply have rewarded length less than information. Testing the first explanation requires deriving both coefficients, not only the length one, from a contextual model and checking that the pair is reproduced; we leave that open. We also note that the survival of the unigram identification for the word term does not establish memorylessness for words: a model with strong sequential dependence can have the same unigram distribution.

\subsection{An illustration on levelled texts}
\begin{table*}[t]
\centering\footnotesize\setlength{\tabcolsep}{4pt}
\caption{OneStopEnglish \citep{vajjala2018}, 189 articles at three levels; level means, each computed from unrounded per-text values. The corpus is used descriptively only; nothing is fitted to it. Per-word surprisal is the mean of $-\log\pi(w)$ over a text's words under the \texttt{wordfreq} distribution; per-sentence surprisal is the text's total word surprisal divided by its number of sentences, averaged over texts.}
\label{tab:ose}
\begin{tabular}{lrrrrrrr}
\toprule
Level & words & $\ASL$ & $\ASW$ & FKGL & FRE & surprisal / word (nat) & surprisal / sentence (nat) \\
\midrule
Elementary & $533$ & $16.95$ & $1.479$ & $8.47$ & $64.5$ & $7.34$ & $124.4$ \\
Intermediate & $672$ & $19.62$ & $1.531$ & $10.13$ & $57.4$ & $7.50$ & $147.3$ \\
Advanced & $828$ & $22.29$ & $1.550$ & $11.40$ & $53.1$ & $7.55$ & $168.3$ \\
\bottomrule
\end{tabular}
\end{table*}

Table~\ref{tab:ose} shows the same discrepancy from the side of the texts. Rewriting a Guardian article from the advanced to the elementary level shortens its sentences by $5.3$ words and lowers the mean unigram surprisal of its words by $0.21\nat$ ($7.55\to7.34$). In Flesch-Kincaid units the resulting difference of $2.93$ grades decomposes into $0.39\times5.34=2.08$ grades from length ($71\%$) and $11.8\times0.071=0.84$ grades from vocabulary ($29\%$). In unigram nats per sentence the same difference, $43.9\nat$, is about $39.7\nat$ from length and $4.1\nat$ from vocabulary ($90\%$ and $10\%$). The formula prices a nat of vocabulary about four times higher than a nat of length here ($0.84/4.1$ against $2.08/39.7$ grades per nat), and the two accountings disagree accordingly. The factor is four rather than the six of Table~\ref{tab:kappa} because the ratio of Table~\ref{tab:kappa} uses the token slope, $2.13\nat$ per syllable, whereas between these level means surprisal rises by $0.21/0.071=2.96\nat$ per syllable; the shortfall is the token-versus-text gap of Section~\ref{sec:data}. Because FKGL and per-sentence surprisal share $\ASL$, their correlation across these texts ($0.91$) is not evidence of anything and we do not use it. The lexical change produced by human simplification is small in information terms, $0.2\nat$ per word, which is consistent with the finding of \citet{gruteke2026} that the classical formulae, FKGL among them, are moderate-to-poor predictors of the reading facilitation produced by exactly this simplification.

\section{The grade step in surprisal, compared with developmental data}
\label{sec:dev}

\begin{figure*}[t]
\centering
\includegraphics[width=\textwidth]{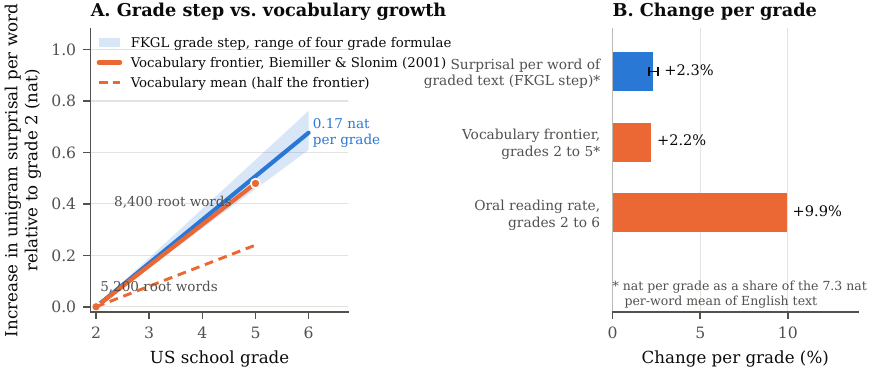}
\caption{The Flesch-Kincaid grade step against developmental data. A: cumulative increase in unigram surprisal per word relative to grade 2 implied by the word terms of the four regression-derived grade formulae (band: $0.15$ to $0.19$ nat per grade; line: their mean, $0.17$), against the log-frequency frontier of the known vocabulary obtained by converting the normative root-word counts of \citet{biemiller2001} through Zipf's law ($\log(8{,}400/5{,}200)=0.48$ nat over three grades) and the mean surprisal of such a vocabulary (half the frontier). B: the same step and the yearly rise of the vocabulary frontier, both expressed as a share of the per-word mean surprisal of English text ($7.3$ nat), which is the scale on which surprisal theory makes them proportional to reading time, beside the yearly change of median oral reading rate \citep{hasbrouck2017}, a percentage of the rate itself. The step is the size of a year of vocabulary growth at the frontier (half that at the vocabulary mean) and several times smaller than a year of fluency growth.}
\label{fig:dev}
\end{figure*}

The Flesch-Kincaid Grade Level rises by one per US school grade by construction, which is why it is the most interpretable of the formulae; the surprisal reading says what that step is in units that the cognitive science of reading uses, and so allows a comparison that the formula on its own does not. We make the comparison for the lexical term, where the reading survives, in two directions; Figure~\ref{fig:dev} summarises it.

\paragraph{What a grade is.} Under the identities, one grade is $1/\kappa_{\mathrm{lex}}=2.7$ to $3.4\nat$ of unigram surprisal per sentence at $L_0=17.6$ words, i.e.\ $0.15$ to $0.19\nat$ per word across the four regression-derived grade formulae ($0.18$ for FKGL), with the factor-of-$1.5$ allowance of Section~\ref{sec:read} (a working allowance from the spread of estimators, not a confidence interval). The conversion holds sentence length fixed, as the partial derivative with respect to $\ASW$ does: it is what the word term charges for a grade. (The length term can also award a grade, $1/0.39=2.6$ extra words per sentence for FKGL, and Sections~\ref{sec:read} and \ref{sec:length} treat it; between the OneStopEnglish levels, which differ mostly in length, mean surprisal rises by $0.072\nat$ per FKGL point, which is the length term at work. No reference length and no levelled corpus enter the per-word conversion itself: $0.18\nat$ is $\lambda/\beta_{\ASW}=2.13/11.8$, the Zipf slope of the language, estimated from the frequency list alone, divided by the formula's own coefficient.) Against a per-word mean of $7.3\nat$ the step is a $2$ to $3\%$ increase per grade, or a $14$ to $17\%$ fall per grade in the geometric-mean frequency of the words a grade is expected to handle. It is a statement about vocabulary, in log frequency, not about speed.

\paragraph{Vocabulary growth.} Children's vocabularies grow by amounts that are reported as counts: the normative estimate of \citet{biemiller2001} puts the mean root-word vocabulary at about $5{,}200$ root words in grade 2 and $8{,}400$ in grade 5, roughly $1{,}000$ root words per year, and \citet{anglin1993} estimates about $1{,}200$ root words per year in the elementary grades, against some $88{,}500$ word families in printed school English \citep{nagy1984}. Counts convert to surprisal through Zipf's rank-frequency law: if the known vocabulary is, to a first approximation, the $V$ most frequent words, the surprisal of the rarest known word is $\log V$ plus a constant, so growth from $V$ to $V+\Delta V$ raises the frontier by $\log(1+\Delta V/V)$. The grade 2 to grade 5 figures give $\log(8{,}400/5{,}200)/3=0.16\nat$ per year; the per-year estimates give $0.12$ to $0.21\nat$ depending on where in the range $V$ is taken. The mean surprisal of a Zipfian vocabulary rises at about half the rate of its frontier. The formula's step, $0.15$ to $0.19\nat$ per word per grade, lies inside the frontier range and within a factor of two of the mean. Which of the two to compare with is a modelling choice: the word term of a formula measures the mean surprisal of a text, and matching it to the frontier says that the texts graded for a year track the words newly learned in that year one for one, whereas matching it to the vocabulary mean says that graded texts sample the whole known vocabulary; the truth is presumably between, and we report both. The conversion from counts to nats uses Zipf's law and assumes that the known vocabulary is approximately a frequency-ordered prefix of the lexicon (order of acquisition correlates with printed frequency, though loosely), that root words can stand for word types, that the Zipf exponent is one in the relevant rank range (an exponent $\alpha$ scales the increment by $\alpha$), and that the population mean count represents the mean of individual log counts; the counts themselves are measurements. It is the first time the two quantities can be put on one axis, and they agree: the formula's step is the size that a year of vocabulary growth produces in log frequency, and it is small because frequency is logarithmic in vocabulary size.

\paragraph{Reading-time development.} Eye-movement studies show that the lexical variables behind surprisal already govern children's reading: eight-year-olds fixate low-frequency words longer than high-frequency words \citep{joseph2013}, length and frequency effects are present in children's silent reading \citep{tiffin2015}, contextual facilitation is present but emerges later in processing than in adults \citep{tiffin2020}, children's eye-movement patterns approach adult ones by about ten to twelve years \citep{blythe2011}, and the development can be modelled as a progressive speeding of lexical processing \citep{reichle2013}. The size of the developmental change is large: in a cross-sectional comparison of Swedish first and second graders reading grade-appropriate (and therefore different) texts, mean fixation duration was $519$ against $349\,\mathrm{ms}$ \citep{strandberg2023}, and median oral reading rates of US children rise from $100$ to $146$ words correct per minute between the ends of grades 2 and 6 \citep{hasbrouck2017}, about $10\%$ per grade. Under surprisal theory, reading time per word is a baseline plus a grade-dependent slope $s_g$ (milliseconds per nat) times surprisal. The formula's step says that the surprisal per word of grade-appropriate text grows, at fixed sentence length, by $2$ to $3\%$ per grade; the fluency norms say that the time per word falls by about $10\%$ per grade, which, if the texts are grade-appropriate, must come from $s_g$ and the baseline. If the linear link held for children's oral reading, which extrapolates a relation established for adult silent reading to a task with decoding, articulation and re-reading components outside it, the comparison would say that most of the yearly change in how fast children read grade-appropriate text is a change in the cost per nat rather than in the information per word; what can be said without the extrapolation is that, on the scale on which surprisal theory makes text information proportional to reading time (nats, linearly), the formula's step is several times smaller than the fluency change. The choice of scale matters and is deliberate: the same year of vocabulary growth is a $17\%$ rise in the number of words known and a $15\%$ fall in the frequency of the frontier word, and on those scales it would not look small; nats are the right scale here because they are the one on which reading time is linear. On that scale the formula's grade is on the vocabulary side of the division. To our knowledge the surprisal slope has not been estimated in children grade by grade with language-model surprisal; the reading makes two predictions for such a study. First, grade-conditional frequency norms \citep[for example the grade-level word counts of][]{zeno1995} should show the mean unigram surprisal of texts levelled at successive grades, compared at equal sentence length and within a genre, rising by about $0.15$ to $0.19\nat$ per word per grade, with the uncertainty above. This prediction has a circularity to guard against: levelled texts have historically been produced under vocabulary lists and under the formulae themselves, so a confirmation on such texts would not separate reading development from publishers' compliance with the formulae, and the test needs texts levelled by an independent criterion, such as children's comprehension. Second, the per-word reading time of children at successive grades should fall by much more than the per-word surprisal of their texts rises, so that the information rate in nats per second grows almost entirely through speed; whether the fall sits in the slope $s_g$ or in the baseline cannot be told from reading rates alone and is a question such a study could answer, as is the relation between the contextual surprisal that reading-time models use and the unigram surprisal converted here. Both predictions are checkable only because the formula's grade has been given units.

\section{Sentence length is not memoryless}
\label{sec:length}

\begin{table*}[t]
\centering\small
\caption{Surprisal of a sentence length $L$ in the Brown corpus ($-\log$ of the empirical proportion of sentences of exactly $L$ words) and under the geometric distribution with the same mean ($q=1/17.62$), in nats.}
\label{tab:length}
\begin{tabular}{lrrrrrrrrrrrr}
\toprule
$L$ & 1 & 2 & 3 & 5 & 8 & 11 & 15 & 20 & 25 & 30 & 40 & 50 \\
\midrule
empirical & $4.28$ & $3.83$ & $3.68$ & $3.40$ & $3.24$ & $3.23$ & $3.30$ & $3.51$ & $3.92$ & $4.30$ & $5.36$ & $6.33$ \\
geometric & $2.87$ & $2.93$ & $2.99$ & $3.10$ & $3.28$ & $3.45$ & $3.69$ & $3.98$ & $4.27$ & $4.56$ & $5.15$ & $5.73$ \\
\bottomrule
\end{tabular}
\end{table*}

The geometric boundary process is the only one under which the sentence-length term of a formula is a surprisal that is linear in $\ASL$. Table~\ref{tab:length} shows that English sentences do not follow it. The surprisal of a sentence length is U-shaped, with its minimum near the modal length of $11$ words: a one-word sentence is as surprising ($4.3\nat$) as a $30$-word sentence. The geometric model overstates the probability of both tails and understates the middle: by a factor of $4.1$ at one word, $0.80$ at eleven, $0.68$ at fifteen, $1.2$ at forty and $1.8$ at fifty. The observed variance ($138.7$) is less than half the geometric variance ($292.8$).

Two consequences follow for the formulae. First, under the empirical length distribution the per-sentence length surprisal is $N_s^{-1}\sum_s-\log P_L(L_s)$, which is not a function of $\ASL$: a text of fifteen-word sentences and a text of alternating five- and twenty-five-word sentences have the same $\ASL$ but length surprisals of $3.30$ and $3.66\nat$ per sentence. A surprisal-faithful length term would need the distribution of lengths within the text, at least its variance, and no classical formula has one. Second, the sentence-length coefficient cannot be read as a boundary surprisal at all (the per-word cost it implies, $1.2\nat$, is twenty times $h_c$, and the boundary term is not linear in length anyway), so its information content, if it has one, must be lexical or structural, as Section~\ref{sec:read} assumed. We do not infer anything about processing difficulty from the U shape: with the word term included, the expected surprisal of a sentence of length $L$ is $HL-\log P_L(L)$, which still increases with $L$ ($7.31+3.83-4.28=6.9\nat$ from one word to two), so the memoryless model is falsified for length as a distribution, not replaced by a prediction that short sentences are hard.

\section{Beyond English: jReadability and categorical features}
\label{sec:ja}

Nothing in Sections~\ref{sec:rep} and~\ref{sec:model} is specific to English or to length counts: the theorem needs an additive word feature, and the identities need a distribution to which the induced one can be compared. Readability formulae of the same shape exist for other languages, and we apply the diagnostic to one that was fitted by regression on levelled corpora and whose word features are categorical rather than lengths. jReadability \citep{hasebe2015,lee2019chapter,lee2016waseda} predicts a six-level readability score for Japanese learner texts,
\begin{equation}
\begin{aligned}
X={}&11.724-0.056\,\ASL-0.126\,r_{\text{kango}}\\&-0.042\,r_{\text{wago}}-0.145\,r_{\text{verb}}-0.044\,r_{\text{part}},
\end{aligned}
\label{eq:jread}
\end{equation}
where $\ASL$ is in UniDic short units per sentence and $r_\bullet$ are percentages of tokens that are Sino-Japanese (\emph{kango}), native (\emph{wago}), verbs and particles; higher $X$ is easier, and the six levels are bands of width one.\footnote{\citet{hasebe2015} define the rates as percentages and the fifth feature as the percentage of particles, as does \citet{lee2016waseda}; \citet{lee2019chapter} give the fifth feature as auxiliary verbs. We analyse both variants; the signs are the same.} The formula was fitted by stepwise regression on $83$ textbooks and balanced-corpus samples levelled by expert judgment, with $R^2=0.896$ \citep{lee2016waseda}. Writing $D=11.724-X$ for difficulty, the lexical term is $N_w^{-1}\sum_i g(w_i)$ with $g(w)=100\,[0.126\,\mathbf 1_{\text{kango}}+0.042\,\mathbf 1_{\text{wago}}+0.145\,\mathbf 1_{\text{verb}}+0.044\,\mathbf 1_{\text{part}}](w)$, so Theorem~\ref{thm:rep} applies with the log-linear model whose sufficient statistics are the four indicators. The identification condition is the multivariate version of the Zipf fit: under a common scale $\kappa$ and reference length $L_0$,
\begin{equation}
\begin{aligned}
\beta_C&=\frac{\kappa L_0}{100}\delta_C,\\ \boldsymbol\delta&=\mathrm{Cov}_\pi(\boldsymbol\phi,\boldsymbol\phi)^{-1}\mathrm{Cov}_\pi(\boldsymbol\phi,-\log\pi),
\end{aligned}
\label{eq:partial}
\end{equation}
where $\boldsymbol\phi$ is the vector of the four indicators and $\boldsymbol\delta$ the vector of \emph{partial} coefficients of the token-weighted regression of unigram surprisal on them, each relative to the classes the indicators leave out (loanwords, hybrids and proper nouns for origin; nouns, adjectives and the other parts of speech for part of speech). Partial coefficients are required because the two attributes are correlated (particles and auxiliaries are native words, loanwords are nouns), so the marginal surprisal of a class need not have the sign of its partial coefficient.

\begin{table*}[t]
\centering\footnotesize\setlength{\tabcolsep}{4pt}
\caption{Japanese word classes under the \texttt{wordfreq} unigram distribution (single UniDic short units, $110{,}049$ types, $96.6\%$ of the list's mass): token mass, marginal mean surprisal (nat), and the partial coefficients $\delta_C$ (nat) of the token-weighted regression of surprisal on the four jReadability indicators, in the variant with particles (part.) and in the variant with auxiliary verbs (aux.). The last column (jR sign) is the sign of the jReadability coefficient in the difficulty direction.}
\label{tab:ja}
\begin{tabular}{llrrrrc}
\toprule
Attribute & Class & mass & marginal & $\delta_C$ (part.) & $\delta_C$ (aux.) & jR sign \\
\midrule
origin & wago & $0.734$ & $5.71$ & $-3.93$ & $-4.60$ & $+$ \\
 & kango & $0.169$ & $9.79$ & $-0.29$ & $-0.28$ & $+$ \\
 & loanword & $0.039$ & $10.77$ & (omitted) & (omitted) & $0$ \\
 & proper noun & $0.027$ & $11.49$ & (omitted) & (omitted) & $0$ \\
\midrule
part of speech & verb & $0.085$ & $8.67$ & $+2.43$ & $+3.10$ & $+$ \\
 & particle & $0.225$ & $3.81$ & $-2.34$ & & $+$ \\
 & auxiliary verb & $0.108$ & $4.72$ & & $-0.75$ & $+$ \\
 & noun & $0.331$ & $9.74$ & (omitted) & (omitted) & $0$ \\
 & adjective & $0.031$ & $7.06$ & (omitted) & (omitted) & $0$ \\
\midrule
$R^2$ of the regression & & & & $0.51$ & $0.44$ & \\
\bottomrule
\end{tabular}
\end{table*}

Table~\ref{tab:ja} gives the partial coefficients. Native words have a partial coefficient of $-3.9$ to $-4.6\nat$: once part of speech is held fixed, a native word is far more predictable than a loanword or a proper noun, and jReadability charges for native words. Particles ($-2.3\nat$) and auxiliary verbs ($-0.75\nat$) are more predictable than the omitted classes, and jReadability charges for them. These two sign reversals are robust to the choice of variant, and they concern the contrasts the formula itself defines, each included class against the classes the formula omits and charges nothing for (loanwords, hybrids and proper nouns for origin; nouns, adjectives and the rest for part of speech). Contrasts between included classes come out differently: native against Sino-Japanese words, and verbs against particles, have the same sign in the formula and in $\boldsymbol\delta$ (in the difficulty direction $-X$, the formula gives $-0.084$ and $+0.101$ per percentage point; $\boldsymbol\delta$ gives $-3.6$ and $+4.8\nat$), so the failure is located in how the formula treats its omitted classes. That suffices: jReadability is, by Theorem~\ref{thm:rep}, the surprisal of \emph{some} distribution, but that distribution is not Japanese. The other two coefficients do not add to the case. The Sino-Japanese partial coefficient is negative ($-0.3\nat$) where jReadability is positive, but by a margin that a different frequency list or dictionary could reverse; and the verb coefficient is positive in both ($+2.4$ to $+3.1\nat$ against the omitted classes, nouns, adjectives and the rest, versus $+0.145$ per percentage point in the formula), so the largest coefficient in the formula is consistent in sign with the Japanese unigram distribution. In sign only: the scale test of Section~\ref{sec:read}, applied to this one coefficient, gives $\kappa_{\mathrm{verb}}=14.5/(\delta_{\mathrm{verb}}L_0)$, which for any plausible reference sentence length $L_0$ (we take $15$ to $30$ short units; jReadability's own length term puts one level at $17.9$) is $0.16$ to $0.40$ levels per nat, against $\kappa_{\mathrm{len}}=0.0081$ below, a ratio of $19$ to $49$ where the English formulae gave $4$ to $8$. The sign test is the weakest of the diagnostics available, and jReadability fails the stronger one more severely than the four regression-derived English grade formulae. We note that a comparison of marginal class means would have given a different and wrong verdict (verbs are less surprising than nouns marginally, $8.67$ against $9.74$); the multivariate condition \eqref{eq:partial} is the one the theorem requires.

What distribution, then? The representation theorem guarantees that one exists and does not single it out. The formula was fitted to level labels, so a distribution conditioned on learner level (or on the textbook level at which a class of words is introduced) is a natural candidate, under which a positive coefficient for native words and particles would mean that these classes grow with level in the levelled corpus faster than the omitted classes; but textbook genre, the correlation between the two attributes, and structural readings (the verb rate times $\ASL/100$ approximates the number of predicates per sentence, and particle and auxiliary chains mark complex predicates) are candidates too, and the surprisal of word classes cannot distinguish them. We state the level-conditional reading as a conjecture to be tested on the levelled corpus, and claim only the negative result.

The length term, $0.056$ per short unit per sentence, puts one level at $17.9$ short units of sentence length; with $H_{\text{ja}}=6.83\nat$ per short unit, and $h_c$ below $0.07\nat$ for any sentence length above $15$ short units, this is $\kappa_{\mathrm{len}}=0.056/(H_{\text{ja}}+h_c)=0.0081$ levels per nat, or about $120\nat$ per level per sentence. The six levels of \eqref{eq:jread} span beginner to advanced and are coarser than US grades, so the number is not comparable to Table~\ref{tab:kappa} directly.

\section{Discussion}
\label{sec:disc}

\paragraph{What the reading buys.} Classical formulae have been attacked for having no theory and defended for working anyway. The surprisal reading gives them a set of conditions rather than a theory: a formula of the additive shape is a faithful memoryless surprisal of the texts it is applied to only to the extent that its word feature predicts unigram surprisal affinely over tokens, its word and length coefficients return the same scale, and the geometric length model it presupposes fits the sentence lengths it is applied to (a cross-entropy is defined whatever the data distribution; faithfulness is the condition). The conditions are checkable, and checking them is the diagnostic this paper proposes. For the English formulae the first condition holds moderately well within a genre and the word coefficients convert, at a Brown-like sentence length, to about three nats per grade per sentence; the second condition fails by a factor of four to eight ($8.3$ for the Navy Farr-Jenkins-Paterson formula); and the third fails outright. The classical formulae are therefore not tangent planes of a memoryless surprisal, but the surprisal reading says precisely how they differ from one, and the scale it extracts from the word term can be compared with anything else that measures reading in information units. Section~\ref{sec:dev} did this for vocabulary growth and reading-rate norms; the same scale can be set against the growth of per-sentence contextual surprisal across graded readers, the Rasch difficulty of cloze items \citep{rasch1960,stenner1996}, or the extra reading time a grade is allowed divided by the reading-time cost of a nat \citep{smith2013,shain2024}. This comparability is the benefit of the reading, and the reason to insist on the units.

\paragraph{What it refutes.} The length term is not a boundary surprisal (twenty times too large for that, and not linear in length) and not a lexical surprisal (four to eight times too small for that), and the sentence-length distribution it presupposes is false. Two of the four lexical terms of jReadability have the wrong sign to be Japanese surprisal. In both cases the representation theorem guarantees that \emph{some} distribution makes the formula a surprisal; the content of the analysis is to show that it is not the obvious one and to list what it could be instead, without claiming that the list has one member.

\paragraph{Readers rather than languages.} Flesch wrote in 1948 that the grade level of children answering test questions was not the best criterion for readability \citep{flesch1948}, and \citet{gruteke2026} have argued that fitting readability measures to comprehension outcomes was the field's original methodological error. The surprisal reading suggests what the criterion could be instead. A readability formula can be read as the cross-entropy of the text under the predictive distribution of a population of readers; its grade scale would then be the rate at which that distribution sharpens with schooling; and the data for fitting it would be estimates of $\pi_r$ (vocabulary-knowledge data, cloze probabilities, or language models adapted to a reader group) rather than comprehension scores. The identities of Proposition~\ref{prop:coef} then become constraints that such data can satisfy or violate, feature by feature.

\paragraph{Which proxies.} Because the identification is only as good as the proxy, the ranking in Table~\ref{tab:constants} matters in practice. As linear predictors of unigram surprisal, the Dale-Chall list and a one-bit mono-/polysyllable indicator are better than the syllable count, and letters are better than syllables; the differences among the best three are small and would need a second frequency source to be trusted. The best proxy, of course, is the surprisal itself, which is why language-model readability measures \citep{collins2004,martinc2021,gruteke2026} outperform the formulae: they drop the proxy.

\paragraph{Large deviations and a language-anchored family.} Remark~\ref{rem:legendre} also points to a way of making the representation non-vacuous without appealing to Zipf's law. Instead of the maximum-entropy family \eqref{eq:gibbs}, which ignores the language, take the family obtained by exponentially tilting the language distribution by the feature, $\pi_\beta(w)=\pi(w)\,e^{\beta f(w)-A_\pi(\beta)}$ with $A_\pi(\beta)=\log\mathbb E_\pi e^{\beta f}$. This family contains the language itself at $\beta=0$, and by Cram\'er's theorem \citep{dembo1998} the probability that a text of $N_w$ words drawn from the language has feature mean $m$ decays as $\exp(-N_w I(m))$ with the rate function
\[
I(m)=\sup_\beta\bigl[\beta m-A_\pi(\beta)\bigr],
\]
the Legendre transform of the cumulant generating function of the feature under the language. $N_w I(\ASW)$ has an exact meaning at finite length: since $\log[P_\beta(T)/P_\pi(T)]=N_w[\beta\,\ASW-A_\pi(\beta)]$, it is the largest log-likelihood gain that a tilt of the language adapted to the text can achieve over the language itself, $\sup_\beta\log[P_\beta(T)/P_\pi(T)]$; by Cram\'er's theorem it is also the exponential rate at which the probability of the text's syllable profile decays, though not that probability's surprisal at finite length. Per word the quantity is $I(\ASW)$, and per sentence it is $\ASL\cdot I(\ASW)$, which brings back the product with $\ASL$ that Section~\ref{sec:model} handled by linearisation; a formula linear in $\ASW$ is then the tangent of $I$ at its calibration point, with slope $\beta(m_0)=I'(m_0)$. This family has a degeneracy that must be stated: $I$ attains its minimum, zero, at the language mean $A_0$, so $I'(A_0)=0$, and a tangent taken at the reference point used throughout this paper has no $\ASW$ slope at all. The tilted reading can reproduce a positive word coefficient only at a calibration point above the language mean, i.e.\ for a grade-dependent reference text, where $\beta(m_0)>0$; at the language mean the comparison with the Zipf identification is trivial ($0$ against $\lambda$), and the Zipf identification is the one that carries the results of Sections~\ref{sec:read} and \ref{sec:dev}. Fitting readability criteria to $I$ with a grade-dependent reference, rather than to a tangent, is a natural next step that we leave open.

\section{Conclusion}
This paper has made two claims. First, the coefficients of the classical readability formulae can be interpreted as surprisal: every formula of the classical additive shape is a sum of cross-entropies under a memoryless model, one per word and one per sentence boundary, and once that model is tied to the language through Zipf's law and the formula is read as its linearisation at a reference text, each coefficient factorises into a scale in grades per nat, a Zipf slope and a reference sentence length. The factorisation is a diagnostic. For English it converts the word terms of the regression-derived grade formulae to about three nats per grade per sentence, shows the length terms to be roughly four to eight times too light for the formulae to be tangent planes of a memoryless surprisal, and shows the sentence-length distribution they presuppose to be false; for Japanese it shows two of jReadability's four lexical coefficients to have the wrong sign. Second, because the Flesch-Kincaid formula outputs grades, and because children's vocabulary is measured by grade in independent research, the interpretation recovers the vocabulary growth the formula implicitly assumes and lets it be checked against development: holding sentence length fixed, the word coefficient converts to $0.18$ nat of unigram surprisal per word per grade ($0.15$ to $0.19$ across the regression-derived formulae), and measured root-word growth, converted through Zipf's law, gives $0.16$ nat per year at the frontier of the known vocabulary. The word coefficient is one year of vocabulary growth, and the formula's grade is a vocabulary scale; reading rate grows several times faster per grade, so the grade does not track fluency. That comparison, and the two predictions for developmental surprisal studies that follow from it, could not be stated while a grade was a number without units; making them statable is what the interpretation is for.

\section*{Limitations}
The word distribution is a general-domain web frequency list of the 2020s, whereas the coefficients were calibrated on texts and readers of 1943 to 1975; the Brown corpus, which is contemporary with the calibrations, gives the same slope, but a reader-specific distribution could differ. The reference length $L_0$ is taken from Brown, $\kappa_{\mathrm{lex}}$ scales as $1/L_0$ (at $L_0=15$ the FKGL scale is $0.37$, at $20$ it is $0.28$), and the calibration passages' typical length is not known. The identities equate regression coefficients with tangent-plane derivatives; relative to the fixed-$\pi$ model the linearisation error has two components (a bilinear cross term and a text-specific residual, Appendix~\ref{app:deriv}), which reach one to two grades for texts far from the reference point and are not absorbed by the formulae' intercepts; the curvature of Remark~\ref{rem:legendre} is a separate quantity, the cost of not re-fitting the reader to each text. The historical coefficients come without standard errors (eighteen passages and $531$ readers for the Navy formulae), so the differences between formulae in Table~\ref{tab:kappa}, informative as their size suggests, cannot be tested for significance. Theorem~\ref{thm:rep} presupposes a finite vocabulary containing every word of the text; the floor for out-of-list words is a scoring convention and does not enter the theorem. The letter slope is used for the stroke counts of the two ARIs without correcting for digits and punctuation. Out-of-list words ($0.4\%$ of OneStopEnglish tokens) are floored at a fixed probability and the renormalisation over the list lowers $H$ slightly; neither affects the token slopes, and moving the floor from $10^{-9}$ to $10^{-8}$ or $10^{-10}$ changes the text-level slopes of Section~\ref{sec:data} by less than $0.01$, the residual standard deviation by $\pm0.007$ and the level means of Table~\ref{tab:ose} by $\pm0.01\nat$. The Dale-Chall list used is the one shipped with \texttt{textstat}, which may differ in detail from the published list, and Gunning's complex-word exclusions are ignored. The OneStopEnglish texts are used descriptively; the three versions of an article are not independent samples. The Japanese analysis uses a frequency list whose entries were segmented with a different dictionary from the one used to classify them, keeps only single-short-unit entries (which drops compounds, disproportionately Sino-Japanese and loanword compounds), and cannot reproduce jReadability's own tokenisation; we expect the native-word and particle results, which are an order of magnitude larger than the Sino-Japanese one, to survive a change of dictionary or frequency source, but we have not verified this, and partial coefficients can move with the covariance among the indicators as well as with the class means. We have not tested the level-conditional reading of jReadability, which requires the levelled corpus.

\section*{Reproducibility}
All constants were computed with \texttt{wordfreq} 3.1.1, \texttt{textstat} 0.7.13, \texttt{cmudict} 1.1.3, \texttt{fugashi} 1.5.2 with \texttt{unidic-lite} 1.0.8, and NLTK 3.10.4 (Brown corpus, Punkt). Scripts that regenerate every table from these packages and the OneStopEnglish corpus accompany the paper.

\section*{Acknowledgements}
This work was supported by JSPS KAKENHI Grant Numbers JP22K12287 and JP26K15085, and by JST, PRESTO Grant Number JPMJPR2363.

\bibliography{refs}

\appendix
\section{Derivations}
\label{app:deriv}

\paragraph{Weighted least squares passes through the means.} With weights $\pi(w)$ summing to one, the minimiser of $\sum_w\pi(w)(I(w)-\lambda\,\syl(w)-\mu)^2$ satisfies $\mu=\sum_w\pi(w)I(w)-\lambda\sum_w\pi(w)\syl(w)=H-\lambda A_0$, hence $\lambda A_0+\mu=H$, which is used in \eqref{eq:ident}; here $A_0$ is the mean syllable count under the weights used in the regression.

\paragraph{Linearisation of \eqref{eq:bilinear} and its error.} Writing $u=\ASL$, $v=\ASW$ and $G(u,v)=\lambda uv+(\mu+h_c)u+(h_\#-h_c)$, the tangent plane at $(L_0,A_0)$ is $G_{\mathrm{tan}}(u,v)=G(L_0,A_0)+(\lambda A_0+\mu+h_c)(u-L_0)+\lambda L_0(v-A_0)$; the coefficients of $u$ and $v$ are those of \eqref{eq:ident}, and the constant $G(L_0,A_0)-(\lambda A_0+\mu+h_c)L_0-\lambda L_0A_0=h_\#-h_c-\lambda L_0A_0$ is absorbed, with $\gamma$, into the formula's intercept. The error of the tangent plane is not absorbed. Writing the token residual $r(w)=-\log\pi(w)-\lambda\syl(w)-\mu$ and its text mean $\bar r_T$, the exact per-sentence surprisal is $\bar I_0(T)=G(u,v)+u\,\bar r_T$, so
\[
\bar I_0(T)-G_{\mathrm{tan}}(u,v)=\lambda\,(u-L_0)(v-A_0)+u\,\bar r_T,
\]
both terms of which vary from text to text. For a primer with $u=6$ and $v=1.25$ the cross term is $2.13\times(6-17.6)\times(1.25-1.47)=5.5\nat$ per sentence, or $1.7$ grades at $\kappa=0.315$; across the OneStopEnglish texts $\bar r_T$ has a standard deviation of $0.23\nat$ per word, i.e.\ $4\nat$ per sentence at $L_0$, or $1.3$ grades.

\paragraph{Indicator features.} For a class $C$ with $\pi(C)=m$ and token-weighted mean surprisals $\bar I_C$, $\bar I_{\bar C}$, the weighted regression of $I$ on $\mathbf 1_C$ has slope $\delta_C=\bar I_C-\bar I_{\bar C}$ and intercept $\bar I_{\bar C}$; the per-sentence surprisal is then $\ASL\,(\delta_C f_C+\bar I_{\bar C})+h_c\ASL+(h_\#-h_c)$ with $f_C$ the fraction of class-$C$ tokens, and its derivative with respect to $f_C$ at $L_0$ is $\delta_C L_0$, giving $\beta_C=\kappa\,\delta_C L_0$ for a coefficient on $f_C$. A coefficient on a percentage is this divided by $100$; a coefficient on the count of class-$C$ words in a fixed sample of $n$ words (as in FORCAST, $n=150$) is this divided by $n$. For several indicators used together, $\delta_C$ is replaced by the partial coefficient of \eqref{eq:partial}.

\paragraph{Coleman-Liau.} The index is $0.0588\,\mathcal L-0.296\,\mathcal S-15.8$ with $\mathcal L$ letters per 100 words and $\mathcal S$ sentences per 100 words, i.e.\ $5.88$ per letter per word and $-29.6/\ASL$; the derivative of the latter at $L_0=17.62$ is $29.6/L_0^2=0.095$ per word per sentence.

\paragraph{Memoryless prediction for the coefficient ratio.} With $\lambda=2.13$, $L_0=17.62$, $H=7.31$ and $h_c=0.058$, $\lambda L_0/(H+h_c)=37.4/7.37=5.1$.

\end{document}